\documentclass{article} 
\usepackage{iclr2017_conference,times}
\usepackage{hyperref}
\usepackage{url}
\usepackage{graphicx}
\usepackage{subfig}
\usepackage{amsmath}
\usepackage{floatrow}
\usepackage{amsfonts}
\usepackage{multirow}
\usepackage{float}
\usepackage{algorithm}
\usepackage{algpseudocode}
\usepackage[innercaption]{sidecap}
\usepackage{pifont}
\newlength{\Oldarrayrulewidth}
\newcommand{\Cline}[2]{%
  \noalign{\global\setlength{\Oldarrayrulewidth}{\arrayrulewidth}}%
  \noalign{\global\setlength{\arrayrulewidth}{#1}}\cline{#2}%
  \noalign{\global\setlength{\arrayrulewidth}{\Oldarrayrulewidth}}}
\title{\centering{Learning a Natural Language Interface \\ with Neural Programmer}}
\author{Arvind Neelakantan\thanks{Work done at Google Brain.} \\
University of Massachusetts Amherst\\
\texttt{arvind@cs.umass.edu} \\
\And
Quoc V. Le \\
Google Brain \\
\texttt{qvl@google.com} \\
\And
Mart{\'{\i}}n Abadi \\
Google Brain \\
\texttt{abadi@google.com} \\
\And
Andrew McCallum\footnotemark[1]  \\
University of Massachusetts Amherst \\
\texttt{mccallum@cs.umass.edu} \\
\And
Dario Amodei\footnotemark[1] \\
{OpenAI} \\
\texttt{damodei@openai.com} \\
}

\iclrfinalcopy 

\begin{document}
\maketitle

\begin{abstract}
Learning a natural language interface for database tables is a challenging task that involves deep language understanding and multi-step reasoning. The task is often approached by mapping natural language queries to {\it logical forms} or {\it programs} that provide the desired response when executed on the database. To our knowledge, this paper presents the first  weakly supervised, end-to-end neural network model to  induce such programs   on a real-world dataset. We enhance the objective function of Neural Programmer, a neural network with built-in discrete operations, and apply it on WikiTableQuestions, a natural language question-answering dataset. The model is trained end-to-end with weak supervision of question-answer pairs, and does not require domain-specific grammars, rules, or annotations that are key elements in previous approaches to program induction.   The main experimental result in this paper is that a single Neural Programmer model achieves 34.2\% accuracy using only 10,000 examples with weak supervision. An ensemble of 15 models, with a trivial combination technique, achieves 37.7\% accuracy, which is competitive to the current state-of-the-art accuracy of 37.1\% obtained by a traditional natural language  semantic parser.  
\end{abstract}

\section{Background and Introduction}

\begin{figure}
  \includegraphics[scale=.35]{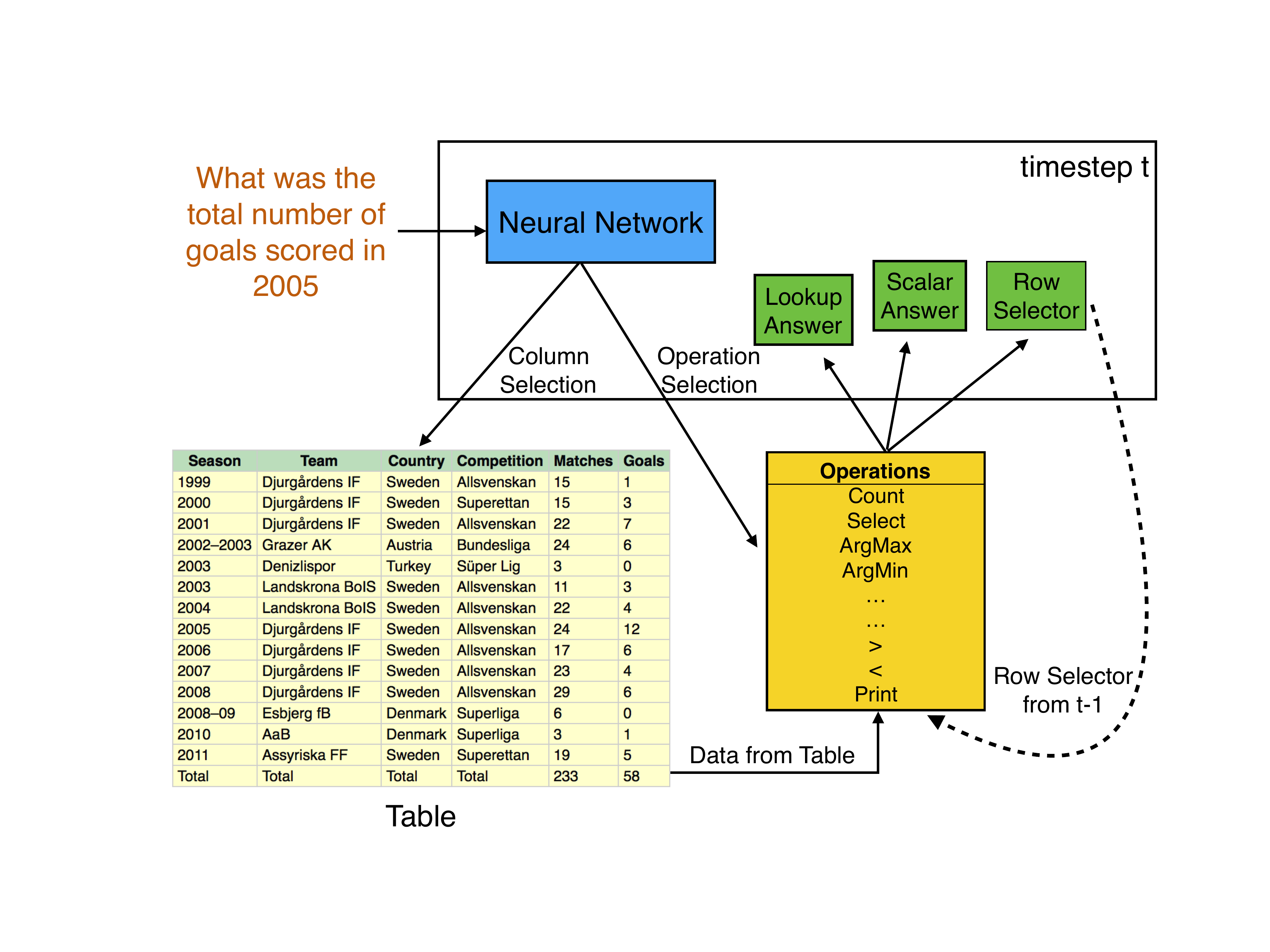}
  \caption{Neural Programmer is a neural network augmented with a set of discrete operations. The model runs for a fixed number of time steps, selecting an operation and a column from the table at every time step. The induced program transfers information across timesteps using the {\it row selector}  variable while the output of the model is stored in the {\it scalar answer} and {\it lookup answer} variables.} \label{intro}
\end{figure} 

Databases are a pervasive way to store and access knowledge. However, it is not straightforward for users to interact with databases since it often requires programming skills and knowledge about database schemas. Overcoming this difficulty by allowing users to communicate with databases via natural language is an active research area. The common approach to this task is by  semantic parsing, which is the process of mapping natural language to symbolic representations of meaning. In this context, semantic parsing yields logical forms or programs that provide the desired response when executed on the databases \citep{zelle:aaai96}.  Semantic parsing is a challenging problem that involves deep language understanding and reasoning with discrete operations such as counting and row selection \citep{percy}.  



The first learning methods for semantic parsing require expensive annotation of question-program pairs \citep{zelle:aaai96,Zettlemoyer05learningto}.  This annotation process is no longer necessary in the  current state-of-the-art semantic parsers that are trained using only question-answer pairs \citep{Liang:2011,kwiatkowski-EtAl:2013:EMNLP,jayant,PasupatL15}.  However, the performance of these methods still heavily depends on domain-specific grammar or pruning strategies to ease  program search. For example, in a recent work on building semantic parsers  for various domains, the authors hand-engineer a separate grammar for each domain \citep{WangBL15}.

Recently, many neural network models have been developed for program induction \citep{jacob,percy-neural,reed2015,zaremba,YinLLK15}, despite the notorious difficulty of handling discrete operations in neural networks~\citep{JoulinM15,neural-gpu}.
Most of these approaches rely on complete programs as supervision \citep{percy-neural,reed2015} while others \citep{zaremba,YinLLK15} have been tried only on synthetic tasks. The work that is most similar to ours is that of \citet{jacob} on the {\it dynamic neural module network}.
However, in their method, the neural network is  employed only to search over a small set of candidate layouts provided by the syntactic parse of the question, and is trained using the REINFORCE algorithm \citep{reinforce}. Hence, their method cannot recover from parser errors, and it is not trivial to adapt the parser to the task at hand. Additionally, all their modules or operations are parametrized by a neural network, so it is difficult to apply their method on tasks that require discrete arithmetic operations.  Finally, their experiments concern a simpler dataset that requires fewer operations, and therefore a smaller search space, than WikiTableQuestions which we consider in our work. We discuss other related work in Section 4.

Neural Programmer \citep{neural-programmer} is  a neural network
augmented with a set of discrete operations. It produces both a program, made up of those operations, and the result
of running the program against a given table.
The operations make use of three variables:
{\it row selector}, {\it scalar answer}, and {\it lookup answer}, which
are updated at every timestep. {\it lookup answer} and {\it scalar
  answer} store answers while {\it row selector} is used to propagate
information across time steps.
As input, a model receives a
question along with a table (Figure \ref{intro}). The model
runs for a fixed number of time steps, selecting an operation and a
column from the table as the argument to the operation at each
time step. During training, soft selection \citep{BahdanauCB14} is
performed so that the model can be trained end-to-end using
backpropagation. This approach allows Neural Programmer to explore the search space
with better sample complexity than hard
selection with the REINFORCE algorithm \citep{reinforce} would provide. All the parameters of the model are learned from a weak
supervision signal that consists of only the final answer; the underlying program,
which consists of a sequence of operations and of selected columns, is latent.


In this work, we develop an approach to semantic parsing based on 
Neural Programmer.  We show how to learn a natural language interface
for answering questions using database tables, thus integrating
differentiable operations that are typical of neural networks with the
declarative knowledge contained in the tables and with discrete operations on tables and entries.  For this
purpose, we make several improvements and adjustments to Neural
Programmer, in particular adapting its objective function to make it more broadly
applicable.

In earlier work, Neural Programmer is applied only on a synthetic
dataset. In that dataset, when the expected answer is an entry in the given table,
its position is explicitly marked in the table.
However, real-world datasets certainly do not include those markers,
and lead to many ambiguities (e.g., \citep{PasupatL15}).
In particular, when the answer is a number that occurs literally in the table, it is not known, a priori,
whether the answer should be generated by an
operation or selected from the table. Similarly, when the answer is a natural language phrase
that occurs in multiple positions in the
table, it is not known which entry (or entries) in the table is actually
responsible for the answer. We extend Neural Programmer
to handle the weaker supervision signal by
backpropagating through decisions that concern how the answer is
generated when there is an ambiguity. 

Our main experimental results concern
WikiTableQuestions~\citep{PasupatL15}, a real-world
question-answering dataset on database tables, with only 10,000 examples for weak
supervision.  This dataset is particularly challenging 
because of its small size and the lack of strong supervision, and also
because the tables provided at test time are never seen during
training, so learning requires adaptation at test time to unseen
column names. A state-of-the-art, traditional  semantic parser that relies on pruning strategies to ease program search achieves 37.1\% accuracy.
Standard neural network models like sequence-to-sequence and pointer networks do not appear to be promising for this dataset, as confirmed in our experiments below, which yield single-digit accuracies. In comparison, a
single Neural Programmer model using minimal text
pre-processing, and trained end-to-end, achieves 34.2\% accuracy.  This surprising  result is enabled primarily by the sample efficiency of Neural Programmer, by  the enhanced objective function,  and by reducing overfitting via strong regularization with dropout \citep{dropout,word-dropout,gal} and weight decay.  An ensemble of
15 models, even with a trivial combination technique, achieves 37.7\%
accuracy.

\section{Neural Programmer}
In this section we describe in greater detail the Neural Programmer model and the modifications we made to the model.  Neural Programmer is a neural network augmented with a set of discrete operations. The model consists of four modules:
\begin{itemize}
\item{{\it Question RNN} that processes the question and converts the tokens to a distributed representation. We use an LSTM network \citep{Hochreiter:1997} as the question RNN.}
\item{A list of discrete operations such as counting and entry selection  that are manually defined. Each operation is parameterized by a real-valued vector that is learned during training.}
\item{A {\it selector} module that induces two probability distributions at every time step, one over the set of operations and another over the set of columns. The input to the selector is obtained by concatenating the last hidden state of the question RNN, the hidden state of the history RNN from the current timestep, and the attention vector obtained by performing soft attention \citep{BahdanauCB14} on the question using the history vector. Following \citet{neural-programmer}, we employ hard selection at test time.} 
\item{{\it History RNN}  modeled by a simple RNN \citep{werbos:bptt} with {\it tanh} activations which remembers the previous operations and columns selected by the model. The input to the history RNN at each timestep is the result of concatenating the weighted representations of operations and columns with their corresponding probability distributions produced by the selector at the previous timestep.}
\end{itemize}
A more detailed description of the basic model can be found in \citet{neural-programmer}. The model runs for fixed total of $T$ timesteps.  The parameters of the operations, selector module, question and history RNNs are all learned with backpropagation using a weak supervision signal that consists of the final answer. Below, we discuss  several modifications to the model to make it more broadly applicable, and easier to train.

\subsection{Operations}
We use $15$ operations in the model that were chosen to closely match the set of operations used in the baseline model \citep{PasupatL15}. All the operations except {\it select} and {\it most frequent entry} operate only on the set of selected rows which is given by the {\it row selector} variable.  Before the first timestep, all the rows in the table are set to be selected. The built-in operations are:
\begin{itemize}
  \item{{\it count} returns the number of selected rows in {\it row selector}.}
  \item{{\it select} and {\it most frequent entry} are operations which are computed only once for every question and output a boolean tensor with size same as the size of the input table. An entry in the output of the {\it select} operation is set to $1$ if the entry matches some phrase in the question. The matched phrases in the question are anonymized to prevent overfitting.  Similarly,  for {\it most frequent entry}, it is set to 1 if the entry is the most frequently occurring one in its column.}
  \item{ {\it argmax}, {\it argmin}, {\it greater than}, {\it less than},  {\it greater than or equal to}, {\it less than or equal to}  are all operations that output a tensor with size same as the size of the input table.}
  \item{  {\it first}, {\it last}, {\it previous} and {\it next}  modify the  {\it row selector}.}
  \item{ {\it print} operation assigns {\it row selector} on the selected column of {\it lookup answer}.}
  \item{{\it reset}  resets {\it row selector} to its initial value. This operation also serves as {\it no-op} when the model needs to induce programs whose complexity is less than $T$.}
\end{itemize}
 All the operations are defined to work with soft selection so that the model can be trained with backpropagation. The operations along with their definitions are discussed in the Appendix.
 
\subsection{Output and Row Selector}
Neural programmer makes use of three variables: {\it row selector},  {\it scalar answer} and {\it lookup answer} which are updated at every timestep. The variable {\it lookup answer} stores answers that are selected from the table while {\it scalar answer} stores numeric answers that are not provided in the table.\footnote{It is possible to extend the model to generate natural language responses using an RNN decoder but it is not the focus of this paper and we leave it for further work.} The induced program transfers information across timesteps using the {\it row selector}  variable which contains rows that are selected by the model.

Given an input table $\Pi$,  containing $M$ rows and $C$ columns ($M$ and $C$ can vary across examples), the output variables at timestep $t$ are given by:
\begin{align*}
  &\mathit{scalar\ answer_t} = \alpha^{op}_t(\mathit{count}) output_t(\mathit{count})  , \\
  &\mathit{lookup\ answer_t}[i][j] = \alpha^{col}_t(j) \alpha^{op}_t(\mathit{print}) \mathit{row\ select_{t-1}}[i], \forall (i, j) i=1,2,\ldots, M, j = 1, 2, \ldots, C
\end{align*}
where $\alpha^{op}_t({\mathit op})$ and $\alpha^{col}_t(j)$ are the probabilities assigned by the selector to operation {\it op} and column {\it j} at timestep $t$ respectively and $output_t(\mathit{count})$ is the output of the count operation at timestep $t$.  The row selector variable at timestep $t$ is obtained by taking the weighted average of the outputs of the remaining operations and is discussed in the Appendix. $\mathit{lookup\ answer_T}[i][j]$ is the probability that the element $(i,j)$ in the input table is in the final answer predicted by the model.

\subsection{Training Objective}
We modify the training objective of Neural Programmer to handle the supervision signal available in real-world settings. In previous work, the position of the answers are explicitly marked in the table when the answer is an entry from the table. However, as discussed in Section 1,   in real-world datasets (e.g., \citep{PasupatL15}) the answer is simply written down introducing two kinds of ambiguities. First, when the answer is a number and if the number is in the table, it is not known whether the loss should be computed using the {\it scalar answer} variable or the {\it lookup answer} variable. Second, when the answer is a natural language phrase and if the phrase occurs in multiple positions in the table, we again do not know which entry (or entries) in the table is actually responsible for generating the answer. We extend Neural Programmer to handle this weaker supervision signal during training by computing the loss only on the prediction that is closest to the desired response.

For scalar answers we compute the square loss: 
\begin{align*}
L_{scalar}(scalar\ answer_T, y) =  \frac{1}{2}(\mathit{scalar\ answer}_{T} - y)^2 \\
\end{align*}
where $y$ is the ground truth answer. We divide $L_{scalar}$ by the number of rows in the input table and do not backpropagate on examples for which the loss is greater than a threshold since it leads to instabilities in training.

When the answer is a list of items $y=(a_1, a_2, \ldots, a_N)$, for each element in the list ($a_i, i =1, 2, \ldots, N$) we compute all the entries in the table that match that element, given by  $S_i = \{(r, c),\ \forall\  (r,c)\ \Pi[r][c] = a_i  \}$. We tackle the ambiguity introduced when an answer item occurs at multiple entries in the table by computing the loss only on the entry which is assigned the highest probability by the model. We construct $g \in \{0,1\}^{M \times C}$, where $g[i, j]$ indicates whether the element $(i, j)$ in the input table is part of the output. We compute log-loss for each entry and the final loss is given by:
\begin{align*}
 L_{lookup}(\mathit{lookup \ answer}_T, y) =  & \sum \limits_{i=1}^{N}  min_{(r, c) \in S_i} (- \log (\mathit{lookup\ answer}_T[r, c]))  \\ & -\frac{1}{MC} \sum \limits_{i=1}^{M} \sum \limits_{j=1}^{C} [g[i,j]\ ==\ 0]\log (1 - \mathit{lookup\ answer}_T[i, j])
\end{align*}
\noindent where $[cond]$ is $1$ when $cond$ is True, and 0 otherwise.

We deal with the ambiguity that occurs when the ground truth is a number and if the number also occurs in the table, by computing the final loss as the {\it soft minimum} of $L_{scalar}$ and $L_{lookup}$. Otherwise, the loss for an example is $L_{scalar}$ when the ground truth is a number and $L_{lookup}$ when the ground truth matches some entries in the table. The two loss functions $L_{scalar}$ and $L_{lookup}$ are in different scales, so we multiply $L_{lookup}$ by a constant factor which we set to 50.0 after a small exploration in our experiments.

Since we employ hard selection at test time, only one among  {\it scalar answer} and {\it lookup answer} is modified at the last timestep. We use the variable that is set at the last timestep as the final output of the model.

\section{Experiments}
We apply Neural Programmer on the WikiTableQuestions dataset \citep{PasupatL15} and compare it to different non-neural baselines including a natural language semantic parser developed by   \citet{PasupatL15}. Further, we also report results from training the sequence-to-sequence model \citep{SutskeverVL14} and a modified version of the pointer networks \citep{pointer-net}.  Our model is implemented in TensorFlow \citep{tensorflow} and the model takes approximately a day to train on a single Tesla K80 GPU. We use double-precision format to store the model parameters since the gradients become undefined values in single-precision format. Our code is available at \url{https://github.com/tensorflow/models/tree/master/neural_programmer}. 

\subsection{Data}
We use the train, development, and test split given by \citet{PasupatL15}. The dataset contains $11321$, $2831$, and $4344$ examples for training, development, and testing respectively. We use their tokenization, number and date pre-processing.  There are examples with answers that are neither number answers nor phrases selected from the table. We ignore these questions during training but the model is penalized during evaluation following \citet{PasupatL15}. The tables provided in the test set are unseen at training, hence requiring the model to adapt to unseen column names at test time. We train only on examples for which the provided table has less than $100$ rows since we run out of GPU memory otherwise, but consider all examples at test time.
\subsection{Training Details} 
We use $T=4$ timesteps in our experiments. Words and operations are represented as $256$ dimensional vectors, and the hidden vectors of the question and the history RNN are also $256$ dimensional. The parameters are initialized uniformly randomly within the range [-0.1, 0.1]. We train the model using the Adam optimizer \citep{KingmaB14} with mini-batches of size $20$. The $\epsilon$ hyperparameter in Adam is set to 1e-6 while others are set to the default values. Since the training set is small compared to other datasets in which neural network models are usually applied, we rely on strong regularization:
\begin{itemize}
\item{We clip the gradients to norm $1$ and employ early-stopping.} 
\item{The occurrences of words that appear less than $10$ times in the training set are replaced by a single unknown word token.}
\item{We add a weight decay penalty with strength $0.0001$.} 
\item{ We use dropout with a keep probability of $0.8$ on input and output vectors of the RNN, and selector, operation and column name  representations \citep{dropout}.}
\item{We use dropout with keep probability of $0.9$ on the recurrent connections of the question RNN and history RNN using the technique from \citet{gal}.}
\item{We use word-dropout \citep{word-dropout} with keep probability of $0.9$. Here, words in the question are randomly replaced with the unknown word token while training.}
\end{itemize}
We tune the dropout rates, regularization strength, and  the $\epsilon$ hyperparameter using grid search on the development data, we fix the other hyperparameters after a small exploration during initial experiments. 

\begin{center}
    \begin{table}
    \begin{tabular}{| c | c | c | c |}
    \hline
    Method & Dev Accuracy & Test Accuracy \\ \hline \hline
    \multicolumn{3}{|c|}{Baselines from \citet{PasupatL15}}\\ \hline 
    Information Retrieval System &13.4 & 12.7 \\ \hline
    Simple Semantic Parser & 23.6 & 24.3 \\ \hline
    Semantic Parser & 37.0 & 37.1 \\ \hline  \hline
    \multicolumn{3}{|c|}{Neural Programmer}\\ \hline 
    Neural Programmer & 34.1 & 34.2 \\ \hline
    Ensemble of 15 Neural Programmer models & 37.5 & 37.7 \\ \hline \hline
    Oracle Score with 15 Neural Programmer models & 50.5 & - \\ \hline
    \end{tabular}
    \caption{Performance of Neural Programmer compared to baselines from \citep{PasupatL15}.  The performance of an ensemble of 15 models is competitive to the current state-of-the-art  natural language  semantic parser.}
    \label{results}
    \end{table}
\end{center}

\subsection{Results}
Table \ref{results} shows the performance of our model in comparison to baselines from \citet{PasupatL15}. The best result from Neural Programmer is achieved by an ensemble of 15 models. The only difference among these models is that the parameters of each model is initialized with a different random seed. We combine the models by averaging the predicted softmax distributions of the models at every timestep. While it is generally believed that neural network models require a large number of training examples compared to simpler linear models to get good performance, our model achieves competitive performance on this small dataset containing only 10,000 examples with weak supervision.

We did not get better results either by using pre-trained word vectors \citep{word2vec} or by pre-training the question RNN with a language modeling objective \citep{dai}. A possible explanation is that the word vectors obtained from unsupervised learning may not be suitable to the task under consideration. For example, the learned representations of words like {\it maximum} and {\it minimum} from unsupervised learning are usually close to each other but for our task it is counterproductive.  We consider replacing soft selection with hard selection and training the model with the REINFORCE algorithm \citep{reinforce}. The model fails to learn in this experiment, probably because the model has to search over millions of symbolic programs for every input question making it highly unlikely to find a program that gives a reward. Hence, the parameters of the model are not updated frequently enough.

\subsubsection{Neural Network Baselines}
To understand the difficulty of the task for neural network models, we experiment with two neural network baselines: the sequence-to-sequence model \citep{SutskeverVL14} and a modified version of the pointer networks \citep{pointer-net}.  The input to the  sequence-to-sequence model is a concatenation of the table and the question, and the decoder produces the output one token at a time. We consider only examples whose input length is less than  $400$ to make the running time reasonable. The resulting dataset has $8,857$ and $1,623$ training and development examples respectively. The accuracy of the best model on this development set after hyperparameter tuning  is only $8.9\%$.  Next, we experiment with pointer networks to select entries in the table as the final answer. We modify pointer networks to have two-attention heads: one to select the column and the other to select entries within a column. Additionally, the model performs multiple pondering steps on the table before returning the final answer. We train this model only on lookup questions, since the model does not have a decoder to generate answers.  We consider only examples whose tables have less than $100$ rows resulting in training and development set consisting of $7,534$ and $1,829$ examples respectively. The accuracy of the best model on this development set after hyperparameter tuning  is only $4.0\%$.
These results confirm our intuition that discrete operations are hard to learn for neural networks particularly with small datasets in real-world settings. 

\subsection{Analysis}

\begin{center}
    \begin{table}
    \begin{tabular}{| c | c |}
    \hline
    Method & Dev Accuracy \\ \hline \hline 
    Neural Programmer & 34.1  \\ \hline
    Neural Programmer - {anonymization} & 33.7  \\ \hline
    Neural Programmer - {match feature} & 31.1 \\ \hline
    Neural Programmer - \{dropout,weight decay\} & 30.3   \\ \hline
    \end{tabular}
    \caption{Model ablation studies. We find that dropout and weight decay, along with the boolean feature indicating a matched table entry for column selection, have a significant effect on the performance of the model.}
    \label{ablation}
    \end{table}
\end{center}

\subsubsection{Model Ablation}
Table \ref{ablation} shows the impact of different model design choices on the final performance. While anonymizing phrases in the question that match some table entry seems to have a small positive effect, regularization has a much larger effect on the performance.  Column selection is performed in \citet{neural-programmer} using only the name of a column; however, this selection procedure is insufficient in real-world settings. For example the column selected in question $3$ in Table \ref{analysis} does not have a corresponding phrase in the question. Hence, to select a column we additionally use a boolean feature that indicates whether an entry in that column matches some phrase in the question. Table \ref{ablation} shows that the addition of this boolean feature has a significant effect on performance.

\begin{center}
    \begin{table}
    \begin{tabular}{| c | c | c | |  c | c | c | c |}
    \hline
  ID  & Question & & Step 1 & Step 2 & Step 3 & Step 4 \\ \hline \hline
  1 &  \multirow{2}{*}{\parbox{4cm}{what is the total number of teams?}} & Operation & - &    -  &   - &   count  \\ \Cline{0.25pt}{3-7}
	&	& Column & -  &     -  &    - &    - \\ \hline \hline
		
  2 &  \multirow{2}{*}{\parbox{4cm}{how many games had more than 1,500 in attendance?}} & Operation  & - &    -  &   $>=$ &   count  \\ \Cline{0.25pt}{3-7}

	&	& Column & -  &     -  &    attendance &    - \\ \hline \hline
		
  3 &  \multirow{3}{*}{\parbox{4cm}{what is the total number of runner-ups listed on the chart?}} & Operation  & - &    -  &   select &   count  \\ \Cline{0.25pt}{3-7}
    	&  &  & & & & \\
	&	& Column & -  &     -  &    outcome &    - \\ \hline  \hline
 
  4 & \multirow{2}{*}{\parbox{4cm}{which year held the most competitions?}} & Operation & - &    -  &   mfe &   print  \\ \Cline{0.25pt}{3-7}
	&	& Column & -  &     -  &    year &    year \\ \hline \hline	
	
  5 & \multirow{2}{*}{\parbox{4cm}{what opponent is listed last on the table?}} & Operation & last &    -  &   last &   print  \\ \Cline{0.25pt}{3-7}
	&	& Column & -  &     -  &    - &    opponent \\ \hline \hline
						
  6 & \multirow{2}{*}{\parbox{4cm}{which section is longest??}} & Operation & - &    -  &   argmax &   print  \\ \Cline{0.25pt}{3-7}
	&	& Column & -  &     -  &    kilometers &    name \\ \hline \hline		   
		
  7 & \multirow{2}{*}{\parbox{4cm}{which engine(s) has the least amount of power?}} & Operation & - &    -  &   argmin &   print  \\ \Cline{0.25pt}{3-7}
	&	& Column & -  &     -  &    power &    engine \\ \hline \hline

  8 & \multirow{2}{*}{\parbox{4cm}{what was claudia roll's time?}} & Operation & - &    -  &   select &   print  \\ \Cline{0.25pt}{3-7}
	&	& Column & -  &     -  &    swimmer &    time \\ \hline \hline		
						
  9 & \multirow{2}{*}{\parbox{4cm}{who had more silver medals, cuba or brazil?}} & Operation & argmax &    select  &   argmax &   print  \\ \Cline{0.25pt}{3-7}
	&	& Column & nation  &     nation  &    silver &    nation \\ \hline \hline

  10 & \multirow{2}{*}{\parbox{4cm}{who was the next appointed director after lee p. brown?}} & Operation & select &    next  &   last &   print  \\ \Cline{0.25pt}{3-7}
	&	& Column & name  &     -  &    - &    name \\ \hline \hline
		
  11 & \multirow{2}{*}{\parbox{4cm}{what team is listed previous to belgium?}} & Operation & select &    previous  &   first &   print  \\ \Cline{0.25pt}{3-7}
	&	& Column & team  &     -  &    - &    team \\ \hline	 
   \end{tabular}
    \caption{A few examples of programs induced by Neural Programmer that generate the correct answer in the development set. mfe is abbreviation for the operation  {\it most frequent entry}. The model runs for $4$ timesteps selecting an operation and a column at every step. The model employs hard selection during evaluation. The column name is displayed in the table only when the operation picked at that step takes in a column as input while the operation is displayed only when it is other than the {\it reset} operation. Programs that choose {\it count} as the final operation produce a number as the final answer while programs that select {\it print} as the final operation produce entries selected from the table as the final answer.
    }
    \label{analysis}
    \end{table}
\end{center}

\subsubsection{Induced Programs}
Table \ref{analysis} shows few  examples  of programs induced by Neural Programmer that yield the correct answer in the development set.  The programs given in Table \ref{analysis} show a  few  characteristics of the learned model. First, our analysis indicates that the model can adapt to unseen column names at test time. For example in Question $3$, the word {\it outcome} occurs only $8$ times in the training set and is replaced with the unknown word token. Second, the model does not always induce the most efficient (with respect to number of operations other than  the {\it reset} operation that are picked) program to solve a task. The last $3$ questions in the table can be solved using simpler programs. Finally, the model does not always induce the correct program to get the ground truth answer. For example, the last $2$ programs will not result in the correct response for all input database tables. The programs would produce the correct response only when the {\it select} operation matches one entry in the table.

\begin{center}
    \begin{table}
    \begin{tabular}{| c | c | c|}
    \hline
    Operation &  Program in Table \ref{analysis} & Amount (\%) \\ \hline \hline
    \multicolumn{3}{|c|}{Scalar Answer}\\ \hline 
    Only Count & 1 & $6.5$ \\ \hline
    Comparison + Count & 2 & $2.1$ \\ \hline 
    Select + Count & 3 & $22.1$ \\ \hline
    Scalar Answer   & 1,2,3 & $30.7$ \\ \hline	\hline
    \multicolumn{3}{|c|}{Lookup Answer}\\ \hline 
    Most Frequent Entry  + Print & 4 & $1.7$ \\ \hline
    First/Last  + Print & 5 & $9.5$ \\ \hline
    Superlative  + Print & 6,7 & $13.5$ \\ \hline
    Select + Print & 8 & $17.5$ \\ \hline
    Select + \{first, last, previous, next, superlative\}  + Print & 9-11 & $27.1$ \\ \hline
    Lookup Answer  & 4-11 & $69.3$ \\ \hline
    \end{tabular}
    \caption{Statistics of the different sequence of operations among the examples answered correctly by the model in the development set. For each sequence of operations in the table, we also  point to corresponding example programs in Table \ref{analysis}. Superlative operations include  {\it argmax} and {\it argmin}, while comparison operations include {\it greater than}, {\it less than},  {\it greater than or equal to} and {\it less than or equal to}. The model induces a program that results in a scalar answer $30.7$\% of the time while the induced program is a table lookup for the remaining questions.  {\it print} and  {\it select} are the two most common operations used $69.3$\% and $66.7$\%  of the time respectively.}
    \label{op}
    \end{table}
\end{center}

\subsubsection{Contribution of Different Operations}
Table \ref{op} shows the contribution of the different operations. The model induces a program that results in a scalar answer $30.7$\% of the time while the induced program is a table lookup for the remaining questions. The two most commonly used operations by the model are {\it print} and  {\it select}.

\subsubsection{Error Analysis}
To conclude this section, we suggest ideas to potentially improve the performance of the model. First, the oracle performance with 15 Neural Programmer models is $50.5$\% on the development set while averaging achieves only $37.5$\% implying that there is still room for improvement. Next, the accuracy of a single model on the training set is $53$\% which is about $20$\% higher than the accuracy in both the development set and the test set. This difference in performance indicates that the model suffers from significant overfitting even after employing strong regularization. It also suggests that the performance of the model could be greatly improved by obtaining more training data.  Nevertheless, there are limits to the performance improvements we may reasonably expect: in particular, as shown in previous work \citep{PasupatL15}, 21\% of questions on a random set of 200 examples in the considered dataset are not answerable because of various issues such as annotation errors and tables requiring advanced normalization.

\section{Other Related Work}
While we discuss in detail various semantic parsing and neural program induction techniques in Section 1, here we briefly describe other relevant work.  Recently, \citet{blunsom} develop a semi-supervised semantic parsing method that uses question-program pairs as supervision.  Concurrently to our work, \citet{neuralsm} propose {\it neural symbolic machine}, a model very similar to Neural Programmer but trained using the REINFORCE algorithm \citep{reinforce}. They use only $2$ discrete operations and run for a total of $3$ timesteps, hence inducing programs that are much simpler  than ours.    Neural networks have also been applied on question-answering datasets that do not require much arithmetic reasoning  \citep{BordesCW14,IyyerBCSD14,sukhbaatar2015weakly,DBLP:journals/corr/PengLLW15,DBLP:journals/corr/HermannKGEKSB15,ama}.  \citet{squad-best} use a neural network model to get state-of-the-art results on a reading comprehension task \citep{squad}.
 
\section{Conclusion}
In this paper, we enhance Neural Programmer to work with weaker supervision signals to make it more broadly applicable. Soft selection during training enables the model to actively explore the space of programs by backpropagation with superior sample complexity. In our experiments, we show that the model achieves performance comparable to a state-of-the-art traditional semantic parser even though the training set contains only 10,000 examples. To our knowledge, this is the first instance of a weakly supervised, end-to-end neural network model that induces programs on a real-world dataset. 

\paragraph{Acknowledgements}
We are grateful to Panupong Pasupat for answering numerous questions about the dataset, and providing pre-processed version of the dataset and the output of the semantic parser.  We thank David Belanger, Samy Bengio, Greg Corrado, Andrew Dai, Jeff Dean, Nando de Freitas, Shixiang Gu, Navdeep Jaitly, Rafal Jozefowicz, Ashish Vaswani, Luke Vilnis, Yuan Yu and Barret Zoph for their suggestions and the Google Brain team for the support. Arvind Neelakantan is supported by a Google PhD fellowship in machine learning.
\bibliography{language_interface}
\bibliographystyle{iclr2017_conference}

\newpage
\section*{Appendix}
\begin{center}
    \begin{table}
     \small
    \begin{tabular}{| c | c | c |}
    \hline
    Type & Operation & Definition \\ \hline \hline
    Aggregate & count & $\mathit{count_t} = \sum\limits_{i=1}^{M} \mathit{row\_select_{t-1}}[i] $ \\ \hline \hline
     {\multirow{2}{*} {Superlative}} & argmax & $max_t[i][j] = \max (0.0, \mathit{row\_select_{t-1}[i ]} - $ \\
    				& & $\sum_{k=1}^{M} ([\Pi[i][j] < \Pi[k][j]] \times \mathit{row\_select_{t-1}[k]})), i = 1, \ldots, M, j = 1, \ldots, C$\\  \Cline{0.25pt}{2-3}
                               & argmin & $min_t[i][j] = \max (0.0, \mathit{row\_select_{t-1}[i ]} - $ \\
    				& & $\sum_{k=1}^{M} ([\Pi[i][j] > \Pi[k][j]]  \times \mathit{row\_select_{t-1}[k]})), i = 1, \ldots, M, j = 1, \ldots, C$ \\ \hline \hline
    {\multirow{4}{*} {Comparison}} & $>$ & $g[i][j] = \Pi[i][j] > \mathit{pivot_{g}}, \forall (i, j), i = 1, \ldots, M, j = 1, \ldots, C$    \\ \Cline{0.25pt}{2-3}
                               & $<$ & $l[i][j] = \Pi[i][j] < \mathit{pivot_{l}}, \forall (i, j), i = 1, \ldots, M, j = 1, \ldots, C$  \\ \Cline{0.25pt}{2-3}
                               & $\geq$ & $ge[i][j] = \Pi[i][j] \geq \mathit{pivot_{ge}}, \forall (i, j), i = 1, \ldots, M, j = 1, \ldots, C$    \\ \Cline{0.25pt}{2-3}
                               &  $\leq$ & $le[i][j] = \Pi[i][j] \leq \mathit{pivot_{le}}, \forall (i, j), i = 1, \ldots, M, j = 1, \ldots, C$  \\ \hline \hline
    {\multirow{6}{*} {Table Ops}} 
    				& select & $s[i][j] = 1.0$ if $\Pi$[i][j] appears in question else 0.0,   \\
    				& & $\forall (i, j), i = 1, \ldots, M, j = 1, \ldots, C$    \\  \Cline{0.25pt}{2-3}
                               & mfe & $mfe[i][j] = 1.0$ if $\Pi$[i][j] is the most common entry in column j else 0.0,   \\
    				& & $\forall (i, j), i = 1, \ldots, M, j = 1, \ldots, C$    \\  \Cline{0.25pt}{2-3}
    				& first & $f_t[i] = \max (0.0, \mathit{row\_select_{t-1}[i ]} - \sum_{j=1}^{i-1} \mathit{row\_select_{t-1}[j]}),$ \\ 
    				& & $i = 1, \ldots, M$\\  \Cline{0.25pt}{2-3}
                               & last & $la_t[i] = \max (0.0, \mathit{row\_select_{t-1}[i ]} - \sum_{j=i+1}^{M} \mathit{row\_select_{t-1}[j]}), $ \\
                               & & $i = 1, \ldots, M$\\ \Cline{0.25pt}{2-3}
                               & previous & $p_t[i] = \mathit{row\_select_{t-1}[i + 1]}, i = 1, \ldots, M - 1$ ;  $p_t[M] = 0$   \\ \Cline{0.25pt}{2-3}
                               & next & $n_t[i] = \mathit{row\_select_{t-1}[i - 1]},  i = 2, \ldots, M$  ;  $n_t[1] = 0$   \\ \hline  \hline                          
    Print & print & $\mathit{lookup\ answer}_t[i][j] = \mathit{row\_select_{t-1}[i]}, \forall (i, j) i=1,\ldots, M, j = 1, \ldots, C$   \\ \hline \hline
    Reset & reset &  $\mathit{r_t[i]} = 1, \forall i = 1, 2, \ldots, M$    \\ \hline 
    \end{tabular}
    \caption{List of all operations provided to the model along with their definitions. mfe is abbreviation for the operation {\it most frequent entry}. $[cond]$ is $1$ when $cond$ is True, and 0 otherwise. Comparison, select, reset and mfe operations are independent of the timestep while all the other operations are computed at every time step. Superlative operations and most frequent entry are computed within a column. The operations calculate the expected output with the respect to the membership probabilities given by the row selector so that they can work with probabilistic selection.}
    \label{operations}
    \end{table}
\end{center}
\subsection*{Operations}
Table \ref{operations} shows the list of operations built into the model along with their definitions. 
\subsection*{Row Selector}
As discussed in Section 2.3, the output variables  {\it scalar answer} and {\it lookup answer}  are calculated using the output of the count operations and print operation respectively. The {\it row selector} is computed using the output of the remaining operations and is given by, 
\begin{align*}
\mathit{row\ selector_t}[i] & =  \sum_{j =1 }^{C} \{ \alpha^{col}_t(j) \alpha^{op}_t(>) \mathit{g}[i][j] + \alpha^{col}_t(j) \alpha^{op}_t(<) \mathit{l}[i][j]  \\& 
							+ \alpha^{col}_t(j) \alpha^{op}_t(\geq) \mathit{ge}[i][j] + \alpha^{col}_t(j) \alpha^{op}_t(\leq) \mathit{le}[i][j], \\&
							+ \alpha^{col}_t(j) \alpha^{op}_t(argmax) \mathit{max_t}[i][j] + \alpha^{col}_t(j) \alpha^{op}_t(argmin_t) \mathit{min}[i][j], \\&
							+ \alpha^{col}_t(j) \alpha^{op}_t(select) \mathit{s}[i][j] + \alpha^{col}_t(j) \alpha^{op}_t(mfe) \mathit{mfe}[i][j] \}  \\& 
							+ \alpha^{op}_t(previous) \mathit{p_{t}}[i] + \alpha^{op}_t(next)  \mathit{n_{t}}[i] +  \alpha^{op}_t(reset) \mathit{r_{t}}[i] \\&
							+ \alpha^{op}_t(first) \mathit{f_{t}}[i] + \alpha^{op}_t(last) \mathit{la_{t}}[i] \\&
							\forall i,  i=1, 2,\ldots, M
\end{align*}
where $\alpha^{op}_t({\mathit op})$ and $\alpha^{col}_t(j)$ are the probabilities assigned by the selector to operation {\it op} and column {\it j} at timestep $t$ respectively.
\end{document}